%% file: main.tex
\documentclass[conference]{IEEEtran}
\IEEEoverridecommandlockouts

\usepackage[
    a4paper,
    left=13.2mm,
    right=13.2mm,
    bottom=42.3mm,
    top=16.7mm,
]{geometry}

\usepackage{graphicx}    % For including images
\usepackage{textcomp}    % For additional text symbols
\usepackage{xcolor}      % For colored text
\usepackage{url}         % For handling URLs
\usepackage{float}       % For [H] placement specifier
\usepackage{caption}     % For customizing captions
\usepackage{subcaption}  % For subfigures
\usepackage{booktabs}    % For professional looking tables
\usepackage{siunitx}     % For better number formatting
\usepackage{multirow}    % For multi-row cells in tables
\usepackage{tabularx}    % For tables with adjustable-width columns
\usepackage{amsmath, amssymb, amsfonts} % For mathematical symbols and environments
\usepackage{algorithm}   % For algorithms
\usepackage{algpseudocode} % For pseudocode in algorithms
\usepackage{pgfplots}    % For creating plots
\usepackage{tikz}        % For drawing graphics
\usetikzlibrary{matrix, positioning, backgrounds, fit}
\usepackage{comment}

\pgfplotsset{compat=1.17}

\usepackage[square, numbers, sort&compress]{natbib} % For citations
\usepackage{hyperref}      % For hyperlinks

\begin{document}

\title{Signal-Centric Remote Sensing via Alternative Preprocessing and Acoustic Processing for ML-Driven Applications}

%\author{\IEEEauthorblockN{Logan Luna, Sirio Jansen-Sanchez, Leo A Ghelarducci, %Ilteris  Demirkiran}\\
%\IEEEauthorblockA{Department of Electrical Engineering and Computer Science, %Embry-Riddle Aeronautical University\\
%Email: lambethl@my.erau.edu, jansenss@my.erau.edu, ghelard@erau.edu, %demir4a4@erau.edu
%}}

%\maketitle

\author{
\IEEEauthorblockN{Logan Luna\thanks{L. Luna is now with the School of Computer Science, College of Computing, Georgia Institute of Technology, Atlanta, GA 30332 USA (e-mail: lluna@gatech.edu).}\thanks{\copyright~2025 IEEE. Personal use of this material is permitted. Permission from IEEE must be obtained for all other uses, in any current or future media, including reprinting/republishing this material for advertising or promotional purposes, creating new collective works, for resale or redistribution to servers or lists, or reuse of any copyrighted component of this work in other works. Published version: DOI 10.1109/SOUTHEASTCON56624.2025.10971547}}
\IEEEauthorblockA{
\textit{Electrical Engineering and Computer Science}\\
\textit{Embry-Riddle Aeronautical University}\\
Daytona Beach, USA}
\and
\IEEEauthorblockN{Sirio Jansen-Sánchez}
\IEEEauthorblockA{
\textit{Electrical Engineering and Computer Science}\\
\textit{Embry-Riddle Aeronautical University}\\
Daytona Beach, USA \\
jansenss@my.erau.edu}
\and
\IEEEauthorblockN{Ilteris Demirkiran}
\IEEEauthorblockA{
\textit{Electrical Engineering and Computer Science}\\
\textit{Embry-Riddle Aeronautical University}\\
Daytona Beach, USA \\
demir4a4@erau.edu}
\and
\IEEEauthorblockN{Leo Ghelarducci}
\IEEEauthorblockA{
\textit{Electrical Engineering and Computer Science}\\
\textit{Embry-Riddle Aeronautical University}\\
Daytona Beach, USA \\
ghelardl@erau.edu}
}

\maketitle
% \textbf{First sonar-based ml paper written listening to radioactive.}
% !!!!

\begin{abstract}

The dominant method of processing sonar data is using image-based representations, requiring the preprocessing of image data on autonomous systems. We propose an alternative data processing method for remote sensing applications via the use of data in Comma-Seperated Value format. Experimentation on our alternative approach shows a reduction of processing time by 91.18\%, an improvement in accurate object detection by Machine Learning, and an increase in SNR (Signal-to-noise ratio), PSNR (Peak signal-to-noise ratio), and other evaluation metrics.
\end{abstract}

\begin{comment}
    \begin{figure}[!htbp]
    \centering
    % Top Image: Raw Sonar Scan
    \begin{subfigure}{1.05\linewidth}
        \centering
        \includegraphics[width=1\linewidth]{Images/GateUnfiltered.jpg}
        \caption{Raw Sonar Scan}
        \label{fig:raw_sonar_scan}
    \end{subfigure}
    %\vspace{0.5cm} % Add vertical spacing between images
    % Bottom Image: Processed Sonar Scan
    % \begin{subfigure}{1.05\linewidth}
    %     \centering
    %     \includegraphics[width=1\linewidth]{Images/CSVGateKMeans.jpg}
    %     \caption{Processed Sonar Scan (Proposed Approach)}
    %     \label{fig:processed_sonar_scan}
    % \end{subfigure}
    \caption{Comparison of raw sonar scan data (top) and processed data using the proposed approach (bottom), which applies median filtering and background subtraction. The processed scan enhances object clarity and reduces noise.}
    %\vspace{-1em}
    \label{fig:sonar_comparison_vertical}
\end{figure}

\end{comment}

% \input{ToDo}

% \input{BrainStorming IDEAS}

\input{Introduction}

\input{RelatedWork}

\input{BackgroundInformation}

\input{Methodology}

\input{Results}

\input{LimitationsAndFutureWork}
\vspace{-0.5em}
\input{Conclusion}

\footnotesize
\vspace{-0.5em}  % Adjust the amount to your liking
\bibliographystyle{IEEEtran}
\bibliography{main}
\end{document}

%% file: Introduction.tex
\section{Introduction} \label{A}

\begin{comment} 
The effectiveness of these technologies hinges on the efficient processing of the acquired data, which is essential for accurate object detection and comprehensive environmental analysis. Sound Navigation and Ranging data processing presents unique challenges, including high levels of noise and the necessity for real-time analysis in dynamic underwater environments. Traditional data processing methods often involve extensive image processing techniques that can be computationally intensive and time-consuming, limiting their applicability in scenarios requiring rapid decision-making. For instance, conventional denoising methods like average filtering and Wiener filtering may not effectively preserve edge details in sonar images, leading to suboptimal detection clarity \cite{sonar_noise_reduction}.
\end{comment}

Visually-based representations have long been established as the state-of-the-art approach to Sonar remote sensing \cite{Williams2007}, underwater navigation \cite{Fairfield2008}, and object detection via sonar \cite{Aykin2016}. Due to this approach, the preprocessing of data output by the sonar into frames must be done onboard autonomous systems \cite{Kim2015}, often being computationally intensive and time-consuming. As a result, this approach is not optimal for scenarios where fast decision-making and object detection are essential. Furthermore, translating the direct sonar data to an image representation can have the effect of losing information during this translation. Lastly, this approach requires denoising methods to be applied on frames rather than the raw output data, and these traditional denoising methods, such as average filtering and Wiener filtering, may not preserve edge details of sonar images effectively, resulting in a poor level of detection clarity \cite{sonar_noise_reduction}.

\begin{comment}   
Recent advancements have explored alternative approaches to enhance sonar data processing. Techniques such as median filtering combined with background subtraction have shown promise in reducing noise while preserving critical features in sonar imagery \cite{sonar_image_enhancement}. Additionally, the implementation of real-time processing systems using Field-Programmable Gate Arrays (FPGAs) has demonstrated significant improvements in processing speed, enabling more efficient data handling in sonar applications \cite{fpga_sonar_processing}.
\end{comment}

Recent work has explored alternative sonar data processing methods. Median filtering and background subtraction, for instance, have been found to be effective at removing noise from sonar imagery while preserving essential features \cite{sonar_image_enhancement}. Another recent development, the use of Field-Programmable Gate Arrays (FPGAs) in the creation of real-time processing systems, has shown a marked increase in the speed of processing to enable better efficiency in handling the data for sonar applications \cite{fpga_sonar_processing}.

\begin{comment}
This research aims to enhance sonar data processing by adopting alternative data handling methods and optimizing processing pipelines. The primary objectives are to improve processing speed and increase the clarity of detected objects, thereby facilitating more efficient and accurate underwater mapping and object detection. By implementing techniques such as handling data in a Comma Separated Value format, median filtering, background subtraction, and optimized data structures, the study seeks to reduce computational overhead and mitigate noise, leading to faster and more reliable data interpretation.
\end{comment}

The objective of this research is to improve sonar data processing by implementing alternative methods in data handling and the optimization of processing pipelines. The primary goals are the improvement of processing speeds and the augmenting of the clarity of detected objects for computationally efficient and accurate underwater mapping and object detection. This study seeks to achieve this by applying handling data in a Comma Separated Value (CSV) format, applying median filtering and background subtraction on CSV data, and employing optimized data structures for faster and more reliable data interpretation.

%% file: RelatedWork.tex
Sonar data has been extensively studied for its application in object detection and environmental mapping, with challenges such as real-time use and noise reduction having already been explored. 

The processing of sonar data for object detection and environmental mapping has been extensively studied, with various techniques proposed to address challenges such as noise reduction and real-time analysis.

\textbf{Noise Reduction Techniques: }
Noise is a significant issue for sonar imagery quality, with filtering methods often necessary, making computationally efficient methods necessary. Hung et al. proposed an effective algorithm for temporal median filtering in background subtraction, leveraging the high correlation between adjacent frames to expedite the median operation \cite{hung2011fast}.

\textbf{Background Subtraction Methods:}
Background subtraction is crucial for isolating objects of interest in sonar images. Traditional methods often struggle with dynamic backgrounds and varying lighting conditions. Piccardi provided a comprehensive review of background subtraction techniques, highlighting the importance of adaptive models that can handle such variations \cite{piccardi2004background}. Additionally, Hung et al.'s fast temporal median filter algorithm enhances background subtraction efficiency by reducing computational complexity \cite{hung2011fast}.

\textbf{Machine Learning in Sonar Object Detection:}
Sivachandra and Kumudham reviewed object detection and classification for sonar imagery. Their findings indicate that deep learning approaches achieve higher accuracy in object recognition on seabeds \cite{sivachandra2024review}. Yang et al. addressed challenges in sonar image object detection, such as widespread noise and lack of high-frequency information, by proposing a Foreground Enhancement Network (FEN). This network incorporates modules for semantic and edge enhancement, improving detection accuracy in noisy sonar images \cite{yang2023foreground}.

\textbf{Automated Machine Learning (AutoML) Approaches:}
These techniques have been applied to streamline the development of sonar object detection models. Kraken Robotics implemented AutoML principles to address challenges such as limited data and class imbalance in underwater object detection. Their approach optimizes the machine learning pipeline, improving detection performance in sonar images \cite{kraken2020object}.

\textbf{Synthetic Data Generation for Training:}
The scarcity of annotated sonar datasets hampers the training of robust deep-learning models. To mitigate this, Preciado-Grijalva et al. developed a pipeline for generating synthetic sonar images using a sonar simulator. This approach facilitates the creation of extensive datasets, enhancing the training process for object detection models \cite{preciado2023sonar}.

\textbf{Real-Time Object Detection Systems:}
Real-time processing is essential for applications like autonomous underwater vehicles. Steadforce developed a real-time object detection solution for sonar data, integrating machine learning models optimized for edge devices. This system assists operators by highlighting relevant details in sonar images, enhancing situational awareness during missions \cite{steadforce2020real}.

%% file: BackgroundInformation.tex
\section{Background}

\begin{figure*}[!htbp]
    \centering
    % First Row
    \begin{subfigure}[b]{0.3\textwidth}
        \centering
        \includegraphics[width=\textwidth]{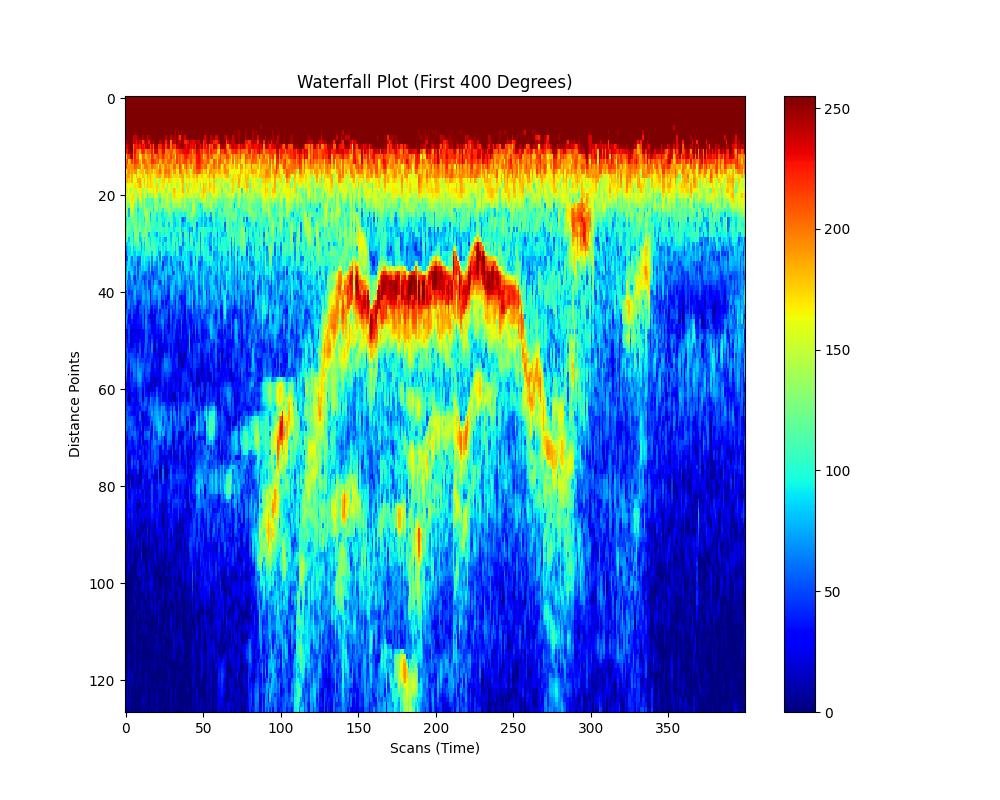}
        \caption{Unfiltered Gate Waterfall Plot}
        \label{fig:GateUnfiltered}
    \end{subfigure}
    \hfill
    \begin{subfigure}[b]{0.3\textwidth}
        \centering
        \includegraphics[width=\textwidth]{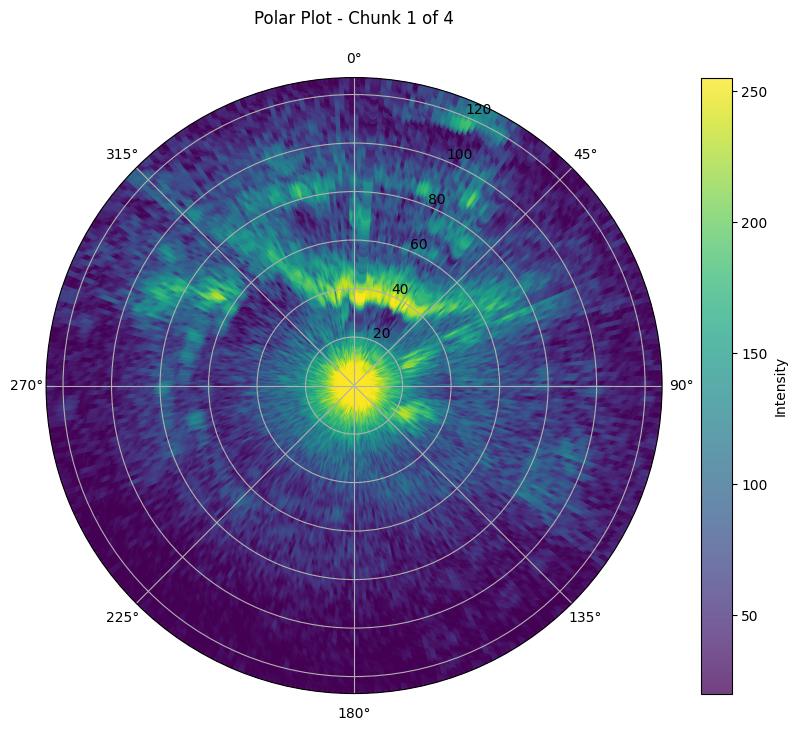}
        \caption{Unfiltered Gate Polar Plot}
        \label{fig:GatePolarPlot}
    \end{subfigure}
    \hfill
    \begin{subfigure}[b]{0.3\textwidth}
        \centering
        \includegraphics[width=\textwidth]{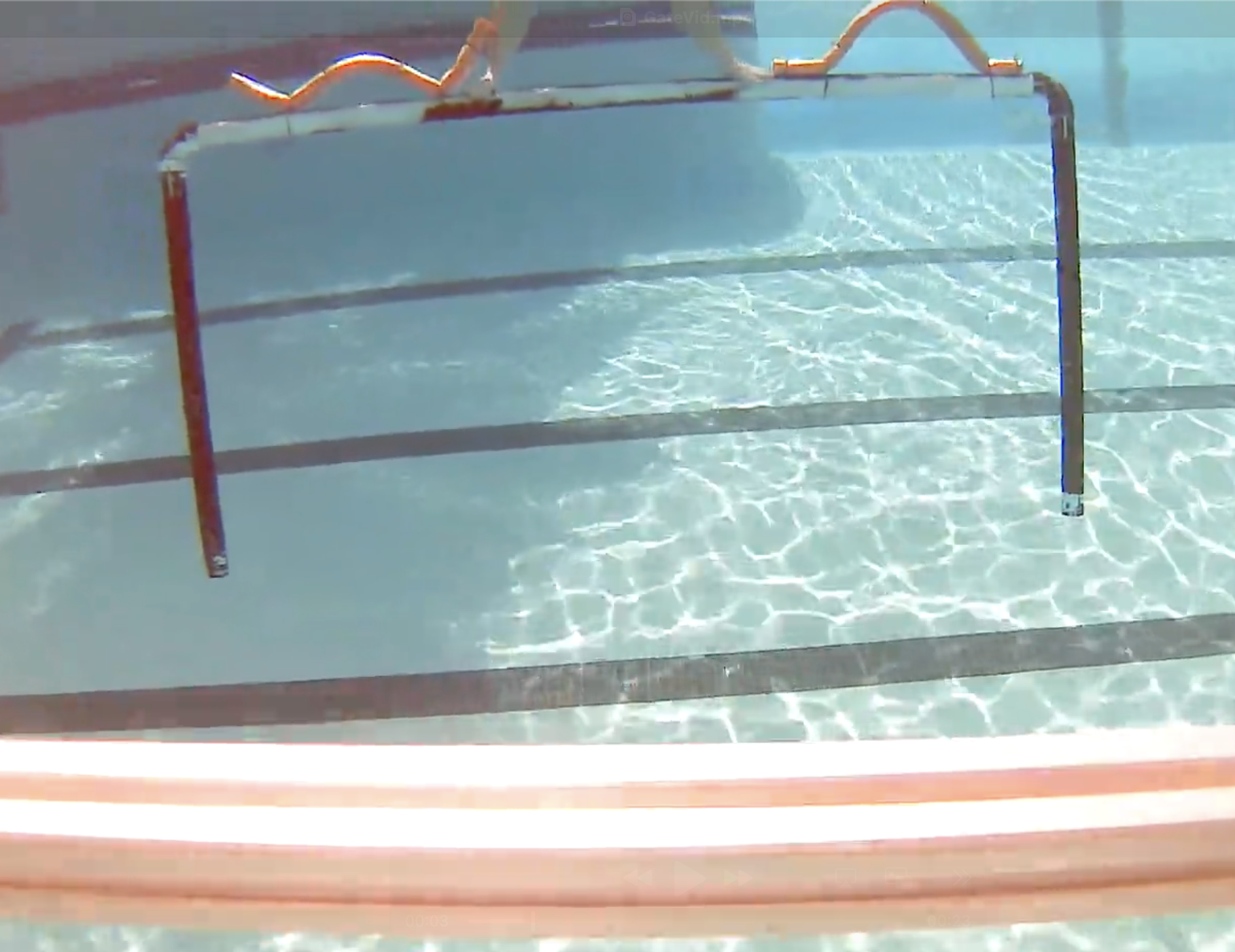}
        \caption{Gate Visual}
        \label{fig:GateVisual}
    \end{subfigure}
    
    \vspace{1em} % Space between rows
    
    % Second Row
    \begin{subfigure}[b]{0.3\textwidth}
        \centering
        \includegraphics[width=\textwidth]{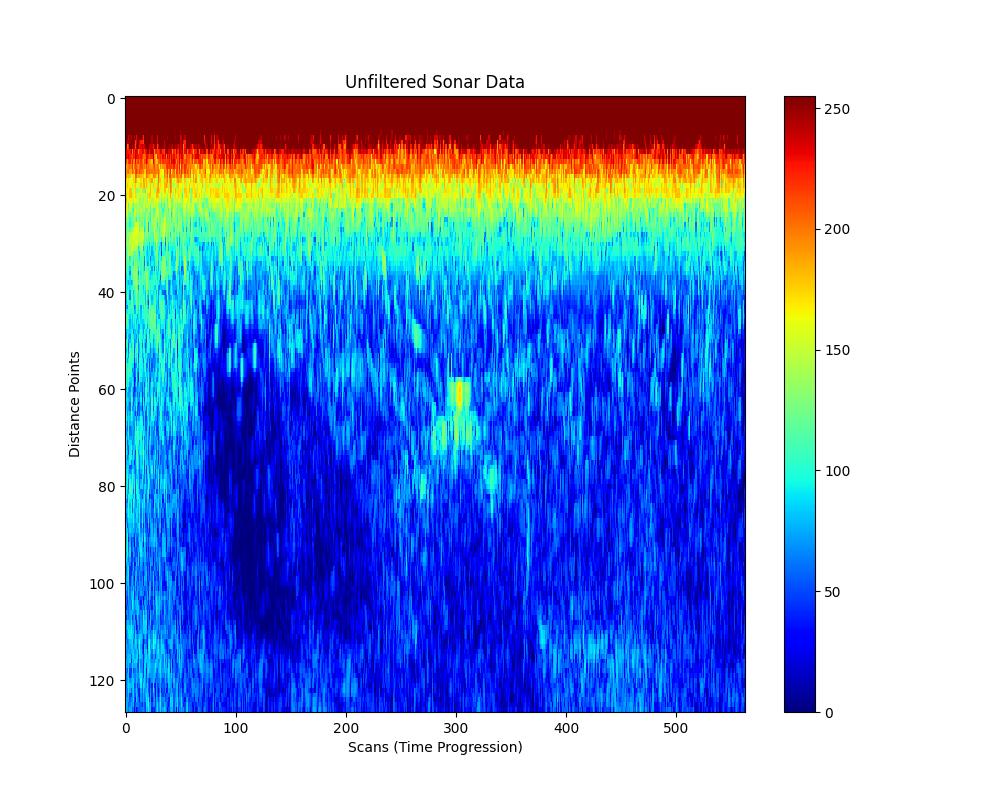}
        \caption{Unfiltered Buoy Waterfall Plot}
        \label{fig:BroomUnfiltered}
    \end{subfigure}
    \hfill
    \begin{subfigure}[b]{0.3\textwidth}
        \centering
        \includegraphics[width=\textwidth]{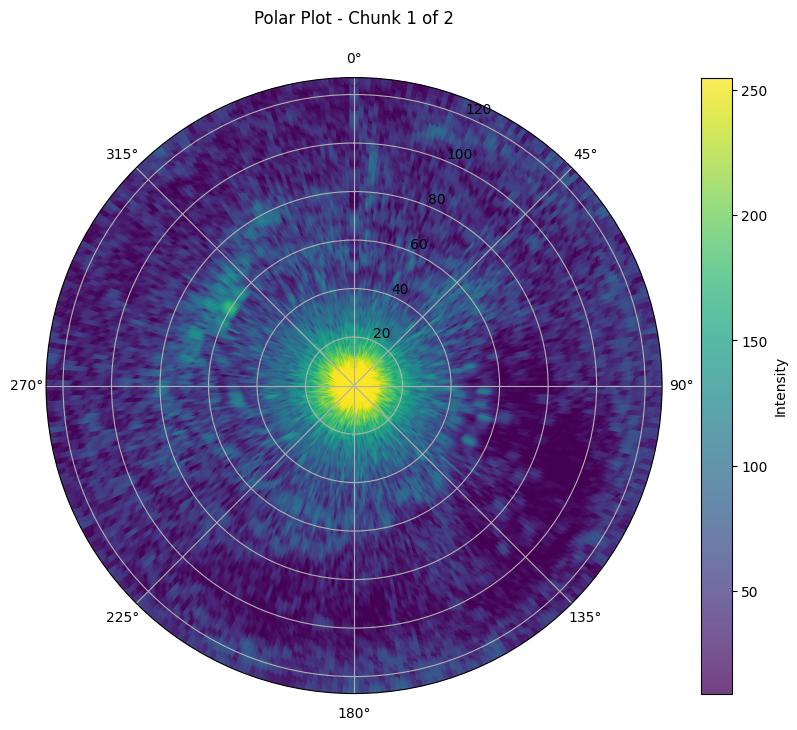}
        \caption{Unfiltered Buoy Polar Plot}
        \label{fig:BroomPolarPlot}
    \end{subfigure}
    \hfill
    \begin{subfigure}[b]{0.3\textwidth}
        \centering
        \includegraphics[width=\textwidth]{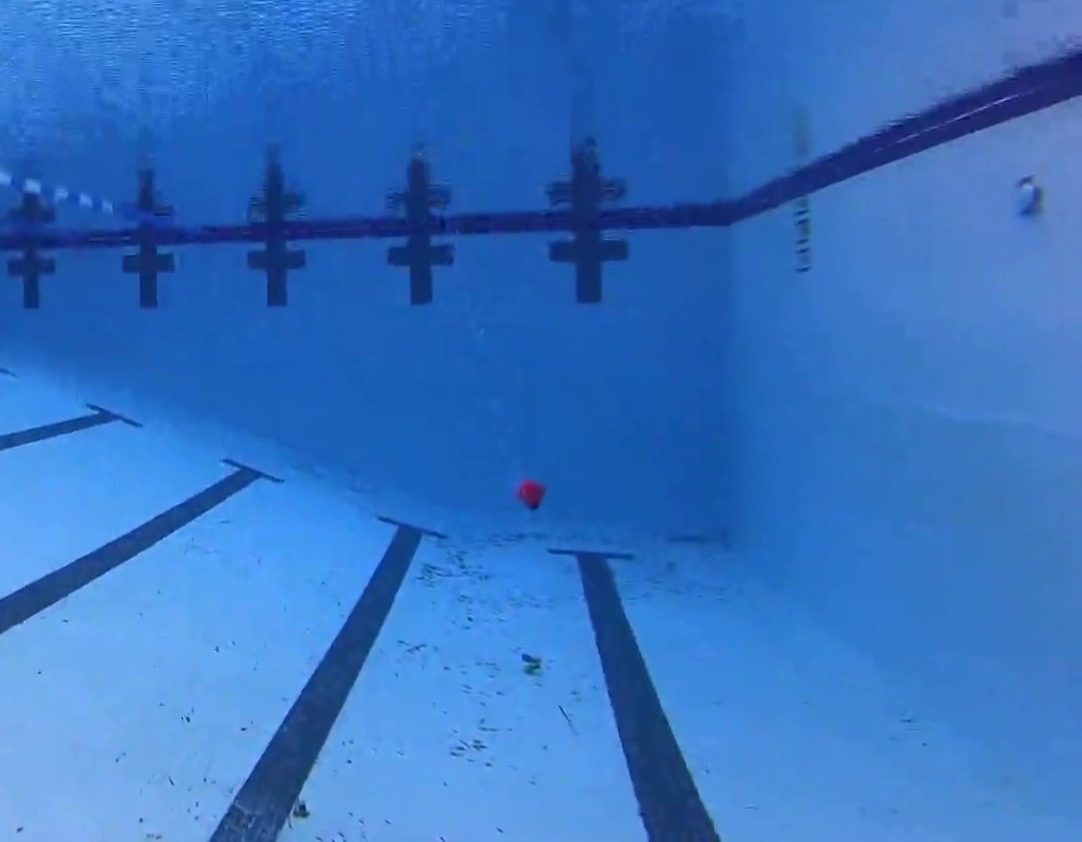}
        \caption{Buoy Visual}
        \label{fig:BroomVisual}
    \end{subfigure}
    
    \caption{Comparison of different views of objects used in the dataset.}
    \label{fig:PolarPlotsCombined}
\end{figure*}

\subsection{Remote Sensing Technologies}

Remote sensing technologies, such as sonar and radar, are indispensable tools for detecting objects and mapping environments across various domains. Sonar utilizes sound waves to navigate, communicate, and detect objects underwater, making it essential for applications such as underwater exploration, navigation, and environmental monitoring. Conversely, radar is crucial for applications like weather forecasting, air traffic control, and military surveillance \cite{RadarVSSonar}.

\subsection{Data Processing in Remote Sensing}

Efficient data processing is paramount in remote sensing to ensure accurate object detection and environmental mapping. Sonar and radar systems generate vast amounts of data that require real-time or near-real-time processing to be actionable. Traditional data processing methods often involve extensive image processing techniques, which can be computationally intensive and may introduce latency, hindering timely decision-making. Additionally, these methods must contend with various sources of noise and interference, such as acoustic reflections and environmental debris in sonar data or atmospheric disturbances in radar data, which can degrade the quality of the detected signals 
\cite{OceanRadar}
.

\subsection{Data Utilized in This Study}

The driver outputs data in CSV format, which is typically processed into visual representations such as waterfall plots (see Figures~\ref{fig:GateUnfiltered} and~\ref{fig:BroomUnfiltered}) and polar plots (see Figures~\ref{fig:GatePolarPlot} and~\ref{fig:BroomPolarPlot}).

A sonar polar plot visualizes sonar data as radial reflections around the sensor, indicating object locations and distances in a circular layout. In contrast, a sonar waterfall plot presents the same data over time as a scrolling 2D chart, highlighting changes in reflections and object movement.

For this study, we will utilize waterfall plots due to their simplicity in visualizing machine learning techniques, such as K-Means clustering, later in the analysis. The objects employed in this study are illustrated below: a larger gate (see Fig. ~\ref{fig:GateVisual}), which is navigated by our testbed Hammerhead during competitions, and a submerged buoy (see Fig. ~\ref{fig:BroomVisual}), used to demonstrate the system's efficiency in detecting smaller objects.

\subsection{Testbed and Hardware Setup}

\begin{figure}[!h]
    \centering
    \includegraphics[width=0.35\textwidth]{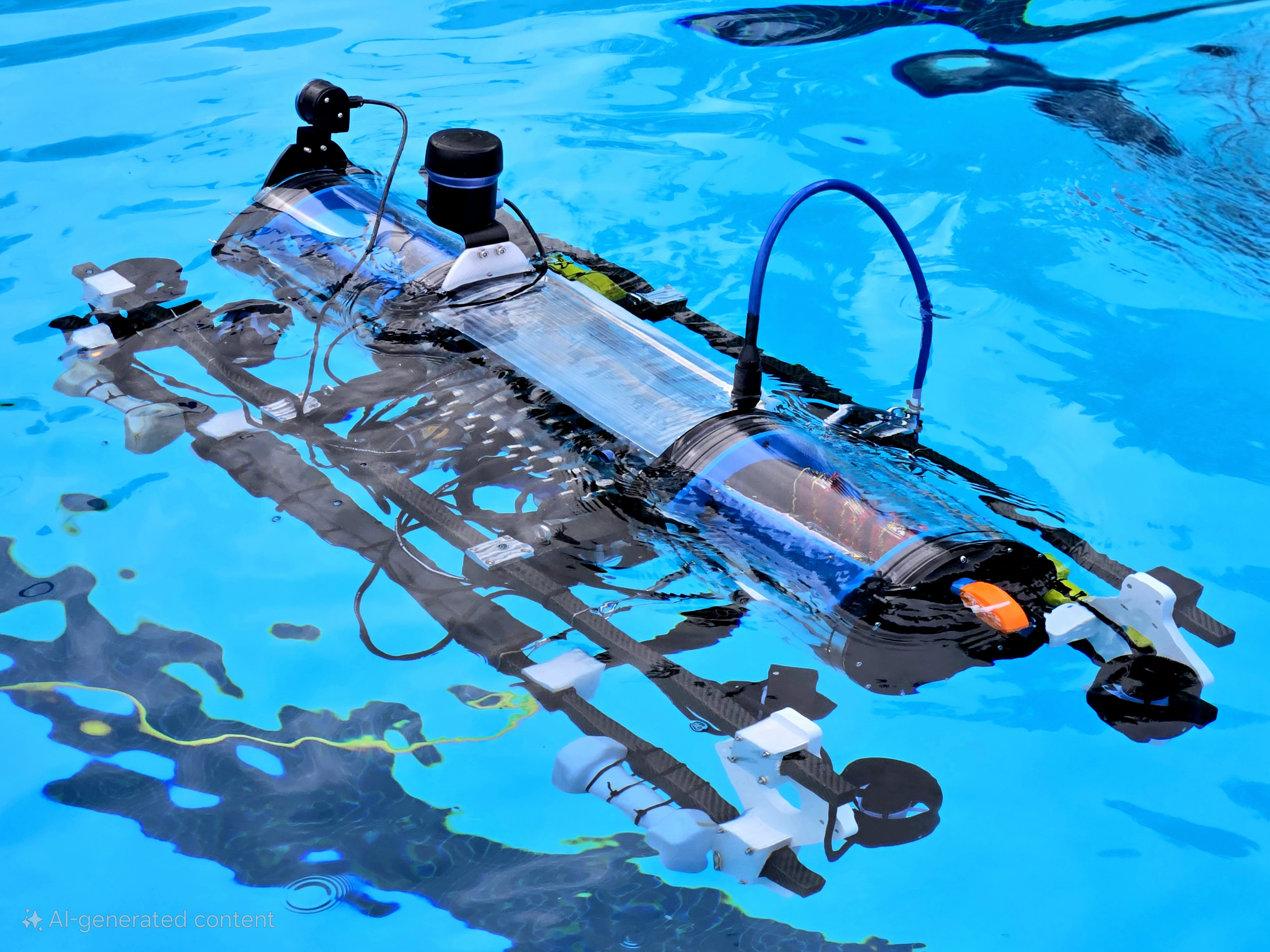}
    \caption{Test Bed}
    \label{fig:Hammerhead}
\end{figure}

The Hammerhead testbed, as shown in Fig. \ref{fig:Hammerhead}, was designed by an undergraduate team. The design revolves around a central connector box, where all external connections terminate, allowing for easy and rapid integration and development of different technologies. The technologies housed in each tube are categorized under two main categories: power and data. Any high current or noisy operations are conducted in separate tubes from those handling data processing, such as a Digital Audio Converter (DAC) for hydrophones or high-frequency digital communications like Ethernet or USB. With the provided thruster configuration, this testbed achieves a full 6 degrees of freedom (DoF) and implements PID loops for attitude and position control.

There is at least one controller per degree of freedom, along with additional controllers for absolute and relative positioning. Feeding the PID control loops there is a Doppler Velocity Logger (DVL) sending instantaneous velocity in the three translational axes, the current dead-reckon position in a Cartesian coordinate system, and a relative orientation. Lastly, there is a Vectonav VN100 Rugged 
%\cite{VN100} 
onboard. This provides the rotational control loops with their orientation as well as our acceleration in each rotational axis. 
This IMU also provides a magnetic orientation in two forms, raw measurement from the sensor and the hard/soft iron compensated measurement. 
The testbed is also equipped with a camera for traditional visual object detection.

\subsubsection{Custom Driver Development}

\begin{figure}[!h]
    \centering
    \includegraphics[width=0.5\textwidth]{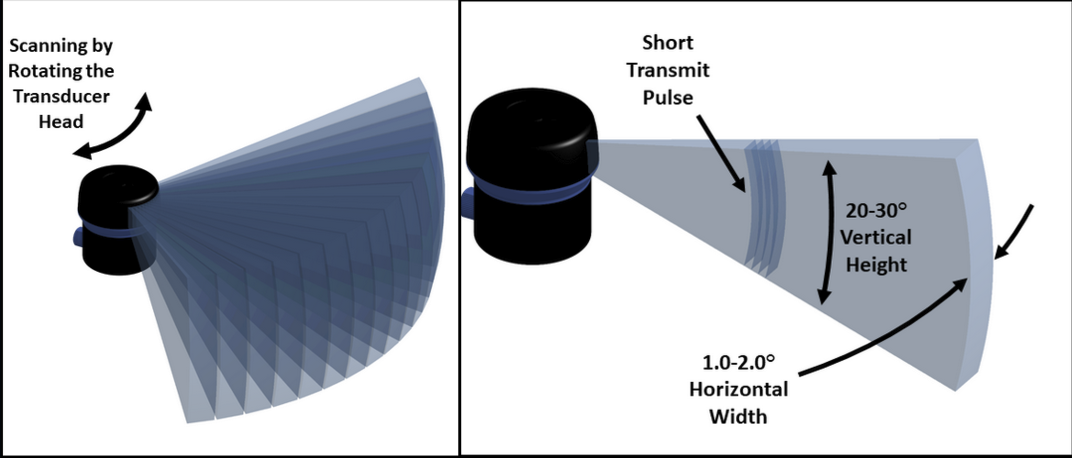}
    \caption{Sonar Hardware Setup \cite{BlueRoboticsSonar}}
    \label{fig:sonar_hardware}
\end{figure}

%The custom driver developed in-house is straightforward.
The custom driver developed in-house consists of a communications thread that opens a TCP socket to the Ping 360 sonar (see Fig. \ref{fig:sonar_hardware}), mounted on the test platform. The sonar communicates using the \textit{BlueRobotics Ping Protocol}.
%\cite{bluerobotics_ping_protocol}. 
The communication thread manages everything required to keep the sonar connected, including sweeping, scanning, and configuring the sonar to current climate conditions.

When combined with the communication middleware, ROS2 Humble, onboard the testbed, it allows for easy integration into the pre-existing control framework. The main interfaces needed to interact with the existing framework include the ability to configure through ROS2, pause and continue continuous sweeps, and scan specific target areas.

\subsection{Hardware}
The main compute unit housed in the Hammerhead testbed is a LattePanda Sigma (LattePanda). The LattePanda is an x86-based computer that consumes 96 watts
%\cite{LattePandaSigma} 
under full load, running Ubuntu Server 22.04. Communications onboard the Hammerhead testbed primarily occur over the onboard gigabit LAN. Some sensors implemented onboard do implement different communication protocols, these are then ingested on a custom board that can translate between the two protocols. 

%% file: Methodology.tex
\section{Methodology}

 \begin{comment}
     
The sonar data utilized in this study comprises multiple sequential scans, each represented as a row in the dataset. Each scan captures a snapshot of the sonar’s readings at a specific timestamp and angle, including:

\begin{itemize}
    \item \textbf{Timestamp}: The precise time when the sonar scan was captured, allowing temporal analysis of object movement and appearance.
    \item \textbf{Angle}: The direction in degrees at which the sonar sensor was pointed during the scan, providing spatial orientation of detected objects.
    \item \textbf{Scan Data}: A space-separated string of intensity values representing sonar reflections at various distances from the sensor. Higher intensity values indicate stronger reflections, likely corresponding to physical objects.

\end{itemize}
\end{comment}

\subsection{Data Structure}
In this study, data collected by this group is utilized and has the following structure. It is comprised of multiple sequential scans, each of an angle from 0 to 400 Gradians represented by a row in the dataset. Each of these scans contains the output data from the sonar at that specific time and angle. The \textbf{UNIX timestamp}, as seen in the first column of Table \ref{tab:sample_data}, is the exact time when the sonar information was captured with nanosecond precision, providing us with the ability of temporal analysis of data. Secondly, the \textbf{angle(°)} in the second column of Table \ref{tab:sample_data} provides us with the direction the sonar is facing when the current scan was taken, enabling the use of spatial orientation for object detection. Lastly, the last column of Table \ref{tab:sample_data} contains the \textbf{scan data} output by the sonar, outputting intensity values representing sonar reflections in a space-separated string. The values of the scan indicate the strength of reflections, with higher values likely corresponding to objects in the environment.

\begin{table}[h]
    \centering
    \caption{Sample Sonar Data Before and After Preprocessing}
    \begin{tabular}{@{}lll@{}}
        \toprule
        \textbf{Timestamp} & \textbf{Angle (°)} & \textbf{Scan Data} \\ \midrule
        1728153122.072070224 & 4 & 255 255 255  161 155 138  60 52 70\\ \bottomrule
    \end{tabular}
    \label{tab:sample_data}
\end{table}

\subsection{Data Preprocessing}

\begin{comment}
    
Data preprocessing is a critical step to ensure the quality and usability of the sonar data for subsequent analysis. The preprocessing pipeline involves the following steps:

\begin{enumerate}
    \item \textbf{Loading and Combining Timestamps}: The raw CSV data contains separate columns for seconds and nanoseconds of the Unix timestamp. These are combined into a single timestamp column to ensure nanosecond precision using Pandas.
    \item \textbf{Cleaning Scan Data}: The scan data columns are merged into a single column, removing any extraneous characters such as ellipses ('...') or additional timestamps. This results in a clean, space-separated string of intensity values.
    \item \textbf{Conversion to Numerical Arrays}: The cleaned scan data strings are converted into numerical arrays using NumPy for efficient processing. This transformation facilitates the application of filtering and clustering algorithms.
\end{enumerate}

\end{comment}

After the output of data by the sonar, the data must be processed before being used for analysis. The preprocessing pipeline is as follows. First, the data is loaded into the pipeline and the separate UNIX timestamps are combined as shown in the first column of Table \ref{tab:OutputData}, maintaining the nanosecond precision. Next, the scan data is merged into a single column and briefly cleaned, removing null spaces, additional timestamps, and other unnecessary characters. Lastly, the scan data strings are converted into Numpy numerical arrays for future efficient processing, with the final output shown in the third column of \ref{tab:OutputData}. This change enables the use of filtering and clustering algorithms on this data.

\begin{table}[h]
    \centering
    \caption{Sample Sonar Data Before and After Preprocessing}
    \begin{tabular}{@{}lll@{}}
        \toprule
        \textbf{Timestamp} & \textbf{Angle (°)} & \textbf{Scan Data} \\ \midrule
        1728153122.072070224 & 4 & 255 255 255  161 155 138  60 52 70\\ \bottomrule
    \end{tabular}
    \label{tab:OutputData}
\end{table}

\begin{comment}

\small
\begin{algorithm}
\caption{Data Preprocessing Pipeline}
\begin{algorithmic}[1]
    \Procedure{PreprocessData}{raw\_csv\_path}
        \State \textbf{Load Data} \Comment{Load CSV data using Pandas}
        \State df $\gets$ \Call{pd.read\_csv}{raw\_csv\_path}
        
        \State \textbf{Combine Timestamps}
        \State df['unix\_timestamp'] $\gets$ df[0].astype(str) + '.' + df[1].astype(str).str.zfill(9)
        \State df.rename(columns={0: 'unix\_timestamp\_sec', 1: 'unix\_timestamp\_ns', 2: 'angle', 3: 'scan\_data'}, inplace=True)
        \State df.drop(columns=['unix\_timestamp\_sec', 'unix\_timestamp\_ns'], inplace=True)
        
        \State \textbf{Clean Scan Data}
        \State df['scan\_data'] $\gets$ \Call{clean\_scan\_data}{df.iloc[:, 2:]}
        
        \State \textbf{Convert to Numerical Arrays}
        \State scan\_matrix $\gets$ \Call{np.array}{df['scan\_data'].tolist()}
        
        \State \Return scan\_matrix
    \EndProcedure
\end{algorithmic}
\end{algorithm}

\end{comment}
\subsection{Proposed Data Handling Methods}

\begin{comment}
To enhance processing speed and object clarity while filtering noise, this study explores alternative data handling and processing approaches. Instead of the conventional image format output by sonar hardware, we propose exporting data in a CSV file format (see Table \ref{tab:sample_data}). This method reduces computational resource consumption and allows for faster processing and more detailed information extraction.
\end{comment}

As described in section \ref{A}, conventional image processing on autonomous underwater vehicle (AUV) systems is a computationally intensive process, especially for continuous image outputs such as sonar readings. To enhance system efficiency, improve object clarity and detection, and more effectively filter noise, this study proposes an alternative data handling and processing approach that circumvents image processing on board AUVs. This process involves changing the data format output of sonar systems to instead be in a CSV file format, as shown in Table \ref{tab:sample_data}. This format is often already used by the sonar, however, the driver will convert this information into image data and this output will be provided. This alternative method reduces computational resource consumption and allows for faster processing and more detailed information extraction.

\subsubsection{Optimized Data Structures and Processing}

To further accelerate the system processing, this study utilizes vectorized operations, parallel processing, and libraries optimized to use large datasets. After using the scan arrays created prior, these NumPy arrays are used to perform vectorized operations on the data to further reduce computation time. 

% \textbf\textbf{TODO, need to improve section above more}

\subsection{Noise Filtering Techniques}

\begin{comment}
    Noise is a significant challenge in remote sensing data, as it can obscure or distort the signals reflected from objects of interest. Effective noise filtering is essential to enhance the signal-to-noise ratio (SNR) and improve the accuracy of object detection. The following techniques are employed to address noise in the data:
\end{comment}

To compare the effectiveness of utilizing CVS data versus Computer Vision (CV) image data, this study will employ the same noise filtering techniques on data processed in both formats and employ machine learning for object detection on this processed data. This approach will provide quantifiable metrics on noise reduction and display the effectiveness of each when paired with machine learning. 

Noise is being addressed in this study as it is a significant challenge in the field of remote sensing. Noise often obscures or distorts singles reflected by points of interest. As a result, effective noise filtering is required for real-time system use. The following techniques are employed to address noise in the data:

\subsubsection{Median Filtering}

\begin{comment}
    Median filtering is a non-linear technique that reduces impulsive noise by replacing each data point with the median of its neighboring values within a specified window size (see Figure \ref{fig:median_filtering}). Unlike linear filtering methods, it effectively preserves edge information, maintaining the integrity of object boundaries while removing spurious noise. This property makes it particularly suitable for accurate object detection in sonar data. The implementation leverages SciPy's optimized \texttt{median\_filter} function for efficient and reliable performance.
\end{comment}

Median filtering is a common technique used in the field of remote sensing, especially sonar. This filtering is a non-linear technique that reduces noise by applying a window of a specified size to replace each data point with the median value of its neighboring values, as displayed in Fig. \ref{fig:median_filtering}. This particular filter effectively maintains data integrity and preserves edge information. The implementation of this filter is done via SciPy's \texttt{median\_filter}.

% Define the input matrix with values ranging from 0 to 255 (7x7)
\newcommand{\inputMatrix}{
    12 \& 45 \& 78 \& 200 \& 34 \& 90 \& 56 \\
    23 \& 67 \& 89 \& 210 \& 123 \& 56 \& 78 \\
    34 \& 56 \& 190 \& 220 \& 145 \& 67 \& 89 \\
    45 \& 78 \& 123 \& 255 \& 167 \& 89 \& 123 \\
    56 \& 90 \& 145 \& 200 \& 189 \& 90 \& 145 \\
    67 \& 123 \& 167 \& 189 \& 200 \& 110 \& 167 \\
    78 \& 145 \& 189 \& 200 \& 210 \& 123 \& 189 \\
}

% Define the median filtered matrix (5x5)
\newcommand{\medianFilteredMatrix}{
    \textbf{45} \& \textbf{89} \& \textbf{123} \& \textbf{167} \& \textbf{145} \\
    \textbf{78} \& \textbf{123} \& \textbf{167} \& \textbf{189} \& \textbf{167} \\
    \textbf{123} \& \textbf{167} \& \textbf{189} \& \textbf{200} \& \textbf{189} \\
    \textbf{145} \& \textbf{189} \& \textbf{200} \& \textbf{210} \& \textbf{200} \\
    \textbf{167} \& \textbf{200} \& \textbf{210} \& \textbf{210} \& \textbf{200} \\
}

% Command to create a median filter visualization
\newcommand{\medianfilterpicture}[3]{% #1: row to be highlighted, #2: column to be highlighted, #3: median value
\begin{tikzpicture}[scale=0.70, transform shape, x=1.5cm, y=1.5cm, >=stealth, font=\sffamily, nodes={align=center}]
    % ------- Styles -------
    \tikzset{%
        parenthesized/.style={%
            left delimiter  = (,
            right delimiter = ),
        },
        node distance = 10mu,
        matrixStyle/.style={
            matrix of nodes,
            nodes in empty cells,
            nodes={minimum size=1cm, draw, anchor=center, text centered},
            row sep=-\pgflinewidth,
            column sep=-\pgflinewidth,
            parenthesized,
            ampersand replacement=\&,
        },
    }

    % ------- Input Matrix -------
    \matrix[matrixStyle] (I) {
        \inputMatrix
    };

    % ------- Equals Symbol -------
    \node (=) [right=of I] {${}={}$};

    % ------- Output Matrix -------
    \matrix[matrixStyle] (O) [right=of {=}] {
        \medianFilteredMatrix
    };

    % ------- Highlighting -------
    \def\rowStart{#1}
    \def\colStart{#2}
    \def\medianValue{#3}

    \newcommand{\padding}{2pt}
    \def\numRowsK{1} % Number of rows in output window (assuming 1 for single cell)
    \def\numColsK{1} % Number of columns in output window (assuming 1 for single cell)

    \newcommand{\windowSize}{3} % 3x3 window

    % Calculate the end row and column for the input window
    \pgfmathtruncatemacro{\rowEnd}{\rowStart + \windowSize -1}
    \pgfmathtruncatemacro{\colEnd}{\colStart + \windowSize -1}

    % Define coordinates for the highlighted input window
    \coordinate (Is-nw) at ([xshift=-\padding, yshift=+\padding] I-\rowStart-\colStart.north west);
    \coordinate (Is-se) at ([xshift=+\padding, yshift=-\padding] I-\rowEnd-\colEnd.south east);
    \coordinate (Is-sw) at (Is-nw |- Is-se);
    \coordinate (Is-ne) at (Is-se |- Is-nw);

    % Fill and draw the red highlighted window in input matrix
    \filldraw[red, fill opacity=.1] (Is-nw) rectangle (Is-se);
    \draw[red, very thick, rounded corners] (Is-nw) rectangle (Is-se);

    % Define coordinates for the highlighted output cell
    \coordinate (Ks-nw) at ([xshift=-\padding, yshift=+\padding] O-\rowStart-\colStart.north west);
    \coordinate (Ks-se) at ([xshift=+\padding, yshift=-\padding] O-\rowStart-\colStart.south east);
    \coordinate (Ks-sw) at (Ks-nw |- Ks-se);
    \coordinate (Ks-ne) at (Ks-se |- Ks-nw);

    % Fill and draw the green highlighted output cell
    \filldraw[green, fill opacity=.1] (Ks-nw) rectangle (Ks-se);
    \draw[green, very thick, rounded corners] (Ks-nw) rectangle (Ks-se);

    % ------- Dotted Lines -------
    \draw[red, dotted, thick] 
        (Is-nw) -- (Ks-nw)
        (Is-se) -- (Ks-se)
        (Is-sw) -- (Ks-sw)
        (Is-ne) -- (Ks-ne)
        ;

    % ------- Median Value Annotation -------
    % \node[above=of (Ks-se)] {Median: \medianValue};

    % ------- Labels -------
    \node[below=of I] (I-label) {$I$ (Input)};
    \node[below=of O] (O-label) {$O$ (Median Filtered Output)};
\end{tikzpicture}
}

% ------- Figure 1: Median Filter -------
\begin{figure}[H]
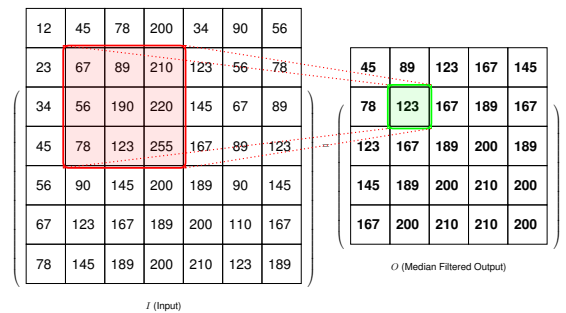

    \centering
    \resizebox{0.4\textwidth}{!}{\medianfilterpicture{2}{2}{123}}
    \caption{Visualization of Median Filtering Process.}
    \label{fig:median_filtering}
\end{figure}

\small
\begin{algorithm}
\caption{Median Filtering}
\begin{algorithmic}[1]
    \Procedure{ApplyMedianFilter}{scan\_matrix, kernel\_size}
        \State \textbf{Initialize} $filtered\_matrix \gets$ a matrix of the same dimensions as $scan\_matrix$
        \State $half\_size \gets \lfloor \frac{kernel\_size}{2} \rfloor$
        \For{$i \gets half\_size$ \textbf{to} $rows - half\_size - 1$}
            \For{$j \gets half\_size$ \textbf{to} $cols - half\_size - 1$}
                \State $window \gets scan\_matrix[i - half\_size : i + half\_size, \; j - half\_size : j + half\_size]$
                \State $median\_value \gets$ \Call{Median}{$window$}
                \State $filtered\_matrix[i, j] \gets median\_value$
            \EndFor
        \EndFor
        \State \textbf{Return} $filtered\_matrix$
    \EndProcedure
\end{algorithmic}
\end{algorithm}

\subsubsection{Background Subtraction}

\begin{comment}
    Background subtraction enhances object detection by removing consistent background signals (see Figure \ref{fig:backgroundsubtraction}). By calculating the median value across scans to estimate the background and subtracting it from each scan, the method isolates anomalies that likely represent objects. Negative values resulting from the subtraction are set to zero to maintain non-negative intensity values, thereby improving the SNR.
\end{comment}

An additional method commonly used in noise reduction is background subtraction. This method removes consistent background information, reducing the amount of potential noise and enhancing object detection. This is achieved by calculating the median value across all scans, estimating the background values, and subtracting these values from scans, as seen in Fig. \ref{fig:backgroundsubtraction}. This effectively isolates anomalies potentially representing objects. This additionally improves SNR by replacing negative values resulting from the subtraction to zero.

% ------- Background Subtraction Picture Command -------
\newcommand{\backgroundsubtractionpicture}[5]{% 
    % #1: row to be highlighted
    % #2: column to be highlighted
    % #3: Input value (I)
    % #4: Background value (B)
    % #5: Output value (O = I - B)
    \begin{tikzpicture}[
        scale=0.50, 
        transform shape, 
        x=0.5cm, 
        y=0.5cm, 
        >=stealth, 
        font=\sffamily, 
        nodes={align=center}
    ]
        % ------- Styles -------
        \tikzset{%
            node distance = 0.5cm, % Reduced node distance for closer spacing
            matrixStyle/.style={
                matrix of nodes,
                nodes in empty cells,
                nodes={minimum size=1cm, draw, anchor=center, text centered},
                row sep=-\pgflinewidth,
                column sep=-\pgflinewidth,
                ampersand replacement=\&
            },
        }

        % ------- Input Matrix -------
        \matrix[matrixStyle] (I) {
            \inputMatrix
        };

        % ------- Minus Sign -------
        \node (minus) [right=of I, xshift=0cm] {$-$}; % Reduced xshift

        % ------- Background Matrix -------
        \matrix[matrixStyle] (B) [right=of minus, xshift=0cm] {
            \backgroundMatrix
        };

        % ------- Equals Symbol -------
        \node (eq) [right=of B, xshift=0cm] {$=$}; % Reduced xshift

        % ------- Output Matrix -------
        \matrix[matrixStyle] (O) [right=of eq, xshift=0cm] { 
            \foregroundMatrix
        };

        % ------- Highlighting -------
        \def\rowStart{#1}
        \def\colStart{#2}
        \def\InputValue{#3}
        \def\BackgroundValue{#4}
        \def\OutputValue{#5}

        \begin{scope}[on background layer]
            \newcommand{\windowSize}{2} % Window size: 2x2

            % Calculate the end row and column for the window
            \pgfmathtruncatemacro{\rowEnd}{\rowStart + \windowSize -1}
            \pgfmathtruncatemacro{\colEnd}{\colStart + \windowSize -1}

            % Highlight the window in the input matrix with a blue background
            \fill[blue!30] 
                (I-\rowStart-\colStart.north west) 
                rectangle 
                (I-\rowEnd-\colEnd.south east);
            \draw[blue, very thick, rounded corners] 
                (I-\rowStart-\colStart.north west) 
                rectangle 
                (I-\rowEnd-\colEnd.south east);

            % Highlight the same window in the background matrix with a yellow background
            \fill[yellow!30] 
                (B-\rowStart-\colStart.north west) 
                rectangle 
                (B-\rowEnd-\colEnd.south east);
            \draw[yellow, very thick, rounded corners] 
                (B-\rowStart-\colStart.north west) 
                rectangle 
                (B-\rowEnd-\colEnd.south east);

            % Highlight the corresponding cell in the output matrix with a green background
            \fill[green!30] 
                (O-\rowStart-\colStart.north west) 
                rectangle 
                (O-\rowStart-\colStart.south east);
            \draw[green, very thick, rounded corners] 
                (O-\rowStart-\colStart.north west) 
                rectangle 
                (O-\rowStart-\colStart.south east);
        \end{scope}

        % ------- Equation -------
        %\node [above=2cm of I.north, align=center] { % Adjusted distance for clarity
        %    \textbf{Mathematical Operation:} \\
        %    For cell (\rowStart,\colStart): \\
        %    $I - B = O$ \\
        %    $ \InputValue - \BackgroundValue = \OutputValue$
        % };

        % ------- Annotations -------
        % Move annotations above the matrices
        \node[above=of I.north, yshift=-0.7cm, align=center] {Input Matrix ($I$)};
        \node[above=of B.north, yshift=-0.7cm, align=center] {Background Matrix ($B$)};
        \node[above=of O.north, yshift=-0.7cm, align=center] {Foreground Output ($O$)};

        % ------- Labels -------
        \node[below=of I] (I-label) {$I$};
        \node[below=of B] (B-label) {$B$};
        \node[below=of O] (O-label) {$O$};

    \end{tikzpicture}
}

% ------- Example Usage -------

% Define the matrices (replace these with your actual matrix data)
\def\inputMatrix{
    10 \& 20 \& 30 \& 40 \\
    50 \& 60 \& 70 \& 80 \\
    90 \& 100 \& 110 \& 120 \\
    130 \& 140 \& 150 \& 160 \\
}

\def\backgroundMatrix{
    5 \& 15 \& 25 \& 35 \\
    45 \& 55 \& 65 \& 75 \\
    85 \& 95 \& 105 \& 115 \\
    125 \& 135 \& 145 \& 155 \\
}

\def\foregroundMatrix{
    5 \& 5 \& 5 \& 5 \\
    5 \& 5 \& 5 \& 5 \\
    5 \& 5 \& 5 \& 5 \\
    5 \& 5 \& 5 \& 5 \\
}

% ------- Figure: Background Subtraction -------
%%
\begin{figure}[htbp]
    \centering
    \resizebox{0.5\textwidth}{!}{\backgroundsubtractionpicture{2}{2}{60}{55}{5}}
    \caption{Background Subtraction Process}
    \label{fig:backgroundsubtraction}
\end{figure}

\begin{comment}
\small
\begin{algorithm}[H]
\caption{Data Preprocessing Pipeline}
\label{alg:data_preprocessing}
\begin{algorithmic}[1]
\REQUIRE 
\begin{itemize}
    \item \texttt{raw\_csv\_path} (string): The file path to the raw CSV data containing sonar scan information.
\end{itemize}
\ENSURE 
\begin{itemize}
    \item \texttt{scan\_matrix} $\in \mathbb{R}^{M \times N}$: A 2D numerical array representing the preprocessed sonar scan data, where \( M \) is the number of distance points and \( N \) is the number of scans.
\end{itemize}

\STATE \textbf{Load Data} \Comment{Load CSV data using Pandas}
\STATE \texttt{df} $\gets$ \Call{pd.read\_csv}{\texttt{raw\_csv\_path}}

\STATE \textbf{Preprocess Scan Data}
\FOR{each row in \texttt{df}}
    \STATE \texttt{df['scan\_data'][row]} $\gets$ \Call{convert\_to\_array}{\texttt{df['scan\_data'][row]}}
\ENDFOR
\COMMENT{Convert scan data from string representation to NumPy arrays}

\STATE \textbf{Extract Scan Matrix}
\STATE \texttt{scan\_matrix} $\gets$ \Call{np.array}{\texttt{df['scan\_data'].tolist()}}

\STATE \Return \texttt{scan\_matrix}
\end{algorithmic}
\end{algorithm}
\end{comment}

\subsection{Object Detection}

\begin{comment}
After preprocessing and filtering, ML techniques are applied to identify and outline detected objects within the sonar data.
\end{comment}

To determine the effectiveness of the described approach and filtering methods, ML for object detection is applied to the filtered CSV data and CV data. This is to assess how each process affects the performance accuracy of  ML applications on this processed data.

\subsubsection{K-Means Clustering and Convex Hull Computation}

\begin{comment}
K-Means clustering groups similar intensity points into clusters, each potentially representing a distinct object. This is followed by computing the convex hull for each significant cluster to outline the detected objects, as illustrated in Figure \ref{fig:K_Means_Process}.

After preprocessing and filtering, K-Means clustering is applied to identify and outline detected objects within the sonar data. This is done across the CSV and CV outputs to compare the performance of this machine learning algorithm based on the data preprocessing method. The clustering process involves the following steps:

\begin{enumerate}
    \item \textbf{Intensity Thresholding}: A stricter intensity threshold is set based on the mean and standard deviation of the intensity values to filter out low-intensity points that are less likely to correspond to significant objects.
    \item \textbf{Clustering}: K-Means clustering is performed on the filtered data to group similar intensity points into clusters, each potentially representing a distinct object, as illustrated by the grouping of colors in Figure \ref{fig:K_Means_After}.
    \item \textbf{Convex Hull Computation}: For each significant cluster (filtered based on size), the convex hull is computed to outline the detected object, providing a clear boundary for visualization. This is visualized by the outlining of groups in Figure \ref{fig:K_Means_After}.
\end{enumerate}

\end{comment}

\begin{figure}[h]
    \centering
    \begin{subfigure}[b]{0.4\textwidth}
        \centering
        \includegraphics[width=\textwidth]{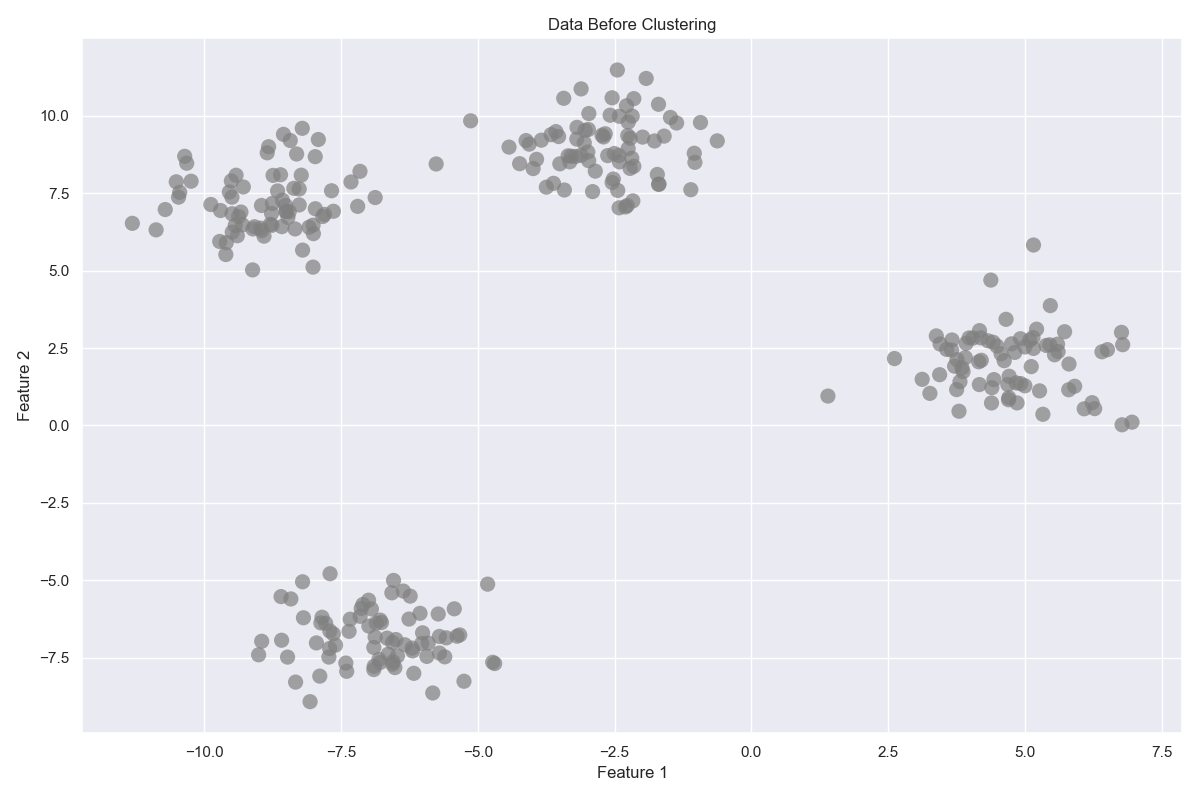}
        \caption{Data Before Being Clustered}
        \label{fig:K_Means_Before}
    \end{subfigure}
    \hfill
    \begin{subfigure}[b]{0.4\textwidth}
        \centering
        \includegraphics[width=\textwidth]{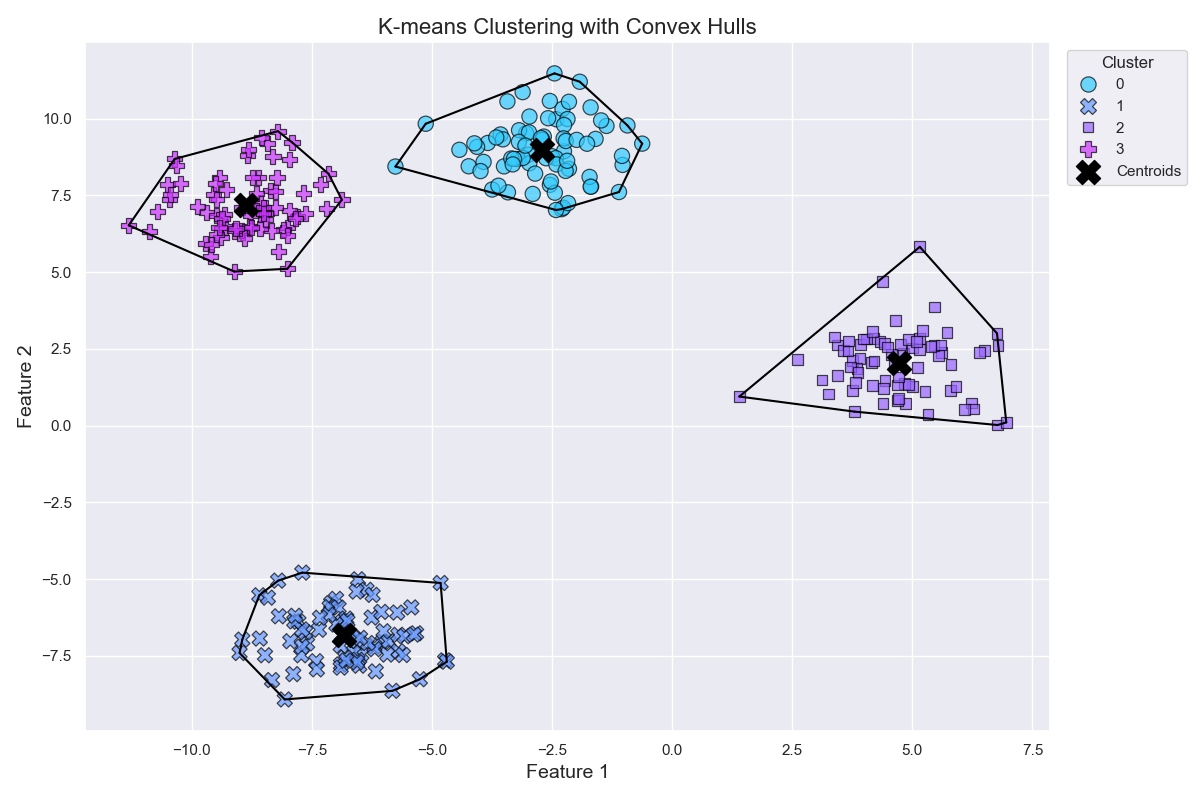}
        \caption{Data After Being Clustered, Outlines Being Convex Hulls}
        \label{fig:K_Means_After}
    \end{subfigure}
    \caption{Display of K-Means Clustering on Sample Data}
    \label{fig:K_Means_Process}
\end{figure}

K-Means performs detections by clustering groups of similar intensity, with each cluster representing a potential object in the environment. Afterward, the convex hull for each cluster is computed to provide an outline of each detected object. This process is illustrated in Fig. \ref{fig:K_Means_Process}.

\small
\begin{algorithm}
\caption{K-Means Clustering and Convex Hull Computation}
\begin{algorithmic}[1]
    \Procedure{DetectObjects}{denoised\_matrix, n\_clusters, intensity\_threshold\_factor, min\_cluster\_size}
        \State \textbf{Flatten Data} to obtain intensity values
        \State Calculate \textbf{Intensity Threshold}:
        \State intensity\_threshold $\gets$ mean(intensity) + intensity\_threshold\_factor $\times$ std(intensity)
        \State \textbf{Filter Points} with intensity > intensity\_threshold
        \State \textbf{Initialize K-Means} with n\_clusters
        \State labels $\gets$ \Call{KMeans.fit\_predict}{filtered\_data}
        
        \State \textbf{Initialize Detections List}
        \State detections $\gets$ []
        
        \For{each unique label in labels}
            \State cluster\_coords $\gets$ filtered\_data[labels == label]
            \If{len(cluster\_coords) < min\_cluster\_size}
                \State \textbf{Skip Small Clusters}
                \State \textbf{continue}
            \EndIf
            \State hull $\gets$ \Call{ConvexHull}{cluster\_coords}
            \State detections.append(\{ 'coords': cluster\_coords[hull.vertices] \})
        \EndFor
        
        \State \Return detections
    \EndProcedure
\end{algorithmic}
\end{algorithm}

\begin{figure*}[!htbp]
    \centering
    % First Row
    \begin{subfigure}[b]{0.3\textwidth}
        \centering
        \includegraphics[width=\textwidth]{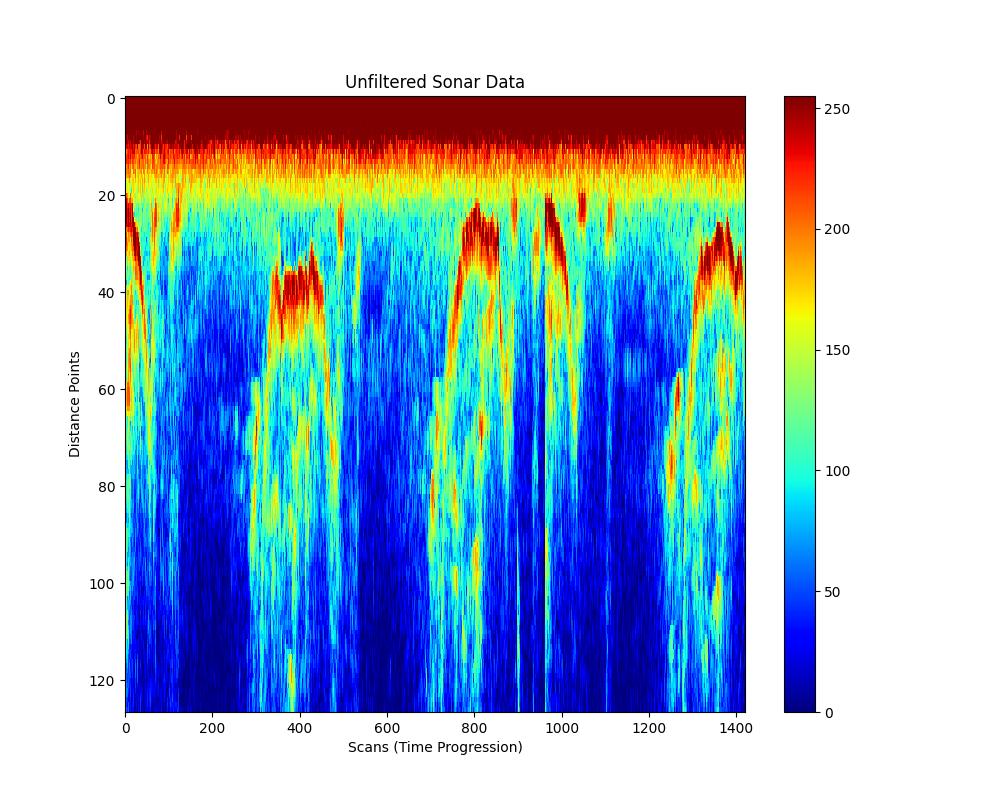}
        \caption{Unfiltered Gate Waterfall Plot}
        \label{fig:original_data}
    \end{subfigure}
    \hfill
    \begin{subfigure}[b]{0.3\textwidth}
        \centering
        \includegraphics[width=\textwidth]{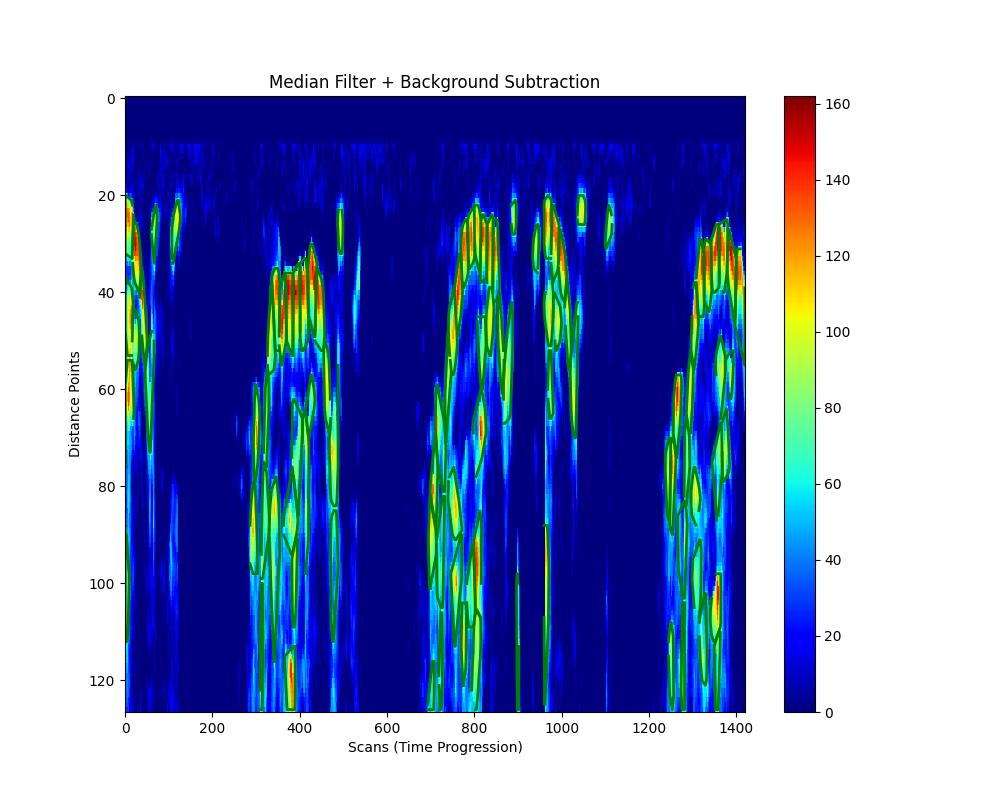}
        \caption{CSV Gate Median Filtering}
        \label{fig:csv_approach}
    \end{subfigure}
    \hfill
    \begin{subfigure}[b]{0.3\textwidth}
        \centering
        \includegraphics[width=\textwidth]{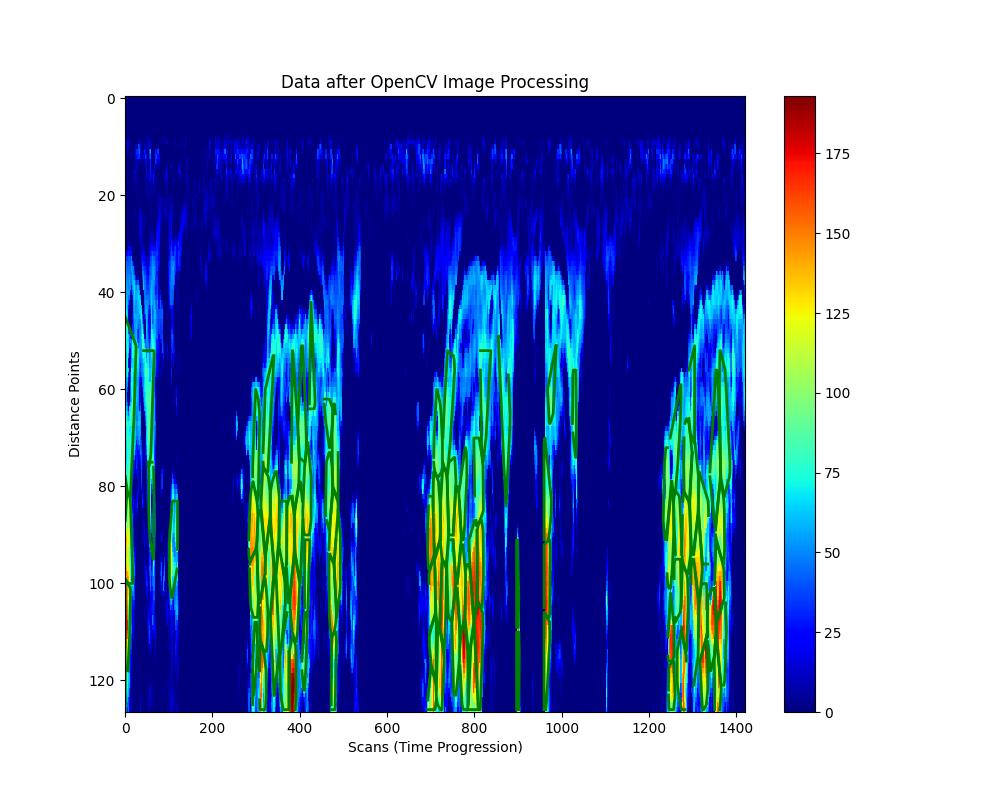}
        \caption{CV Gate Filtering}
        \label{fig:cv_approach}
    \end{subfigure}
    
    \vspace{1em} % Space between rows
    
    % Second Row
    \begin{subfigure}[b]{0.3\textwidth}
        \centering
        \includegraphics[width=\textwidth]{Images/BroomUnfiltered.jpg}
        \caption{Unfiltered Broom Waterfall Plot}
        \label{fig:Broom}
    \end{subfigure}
    \hfill
    \begin{subfigure}[b]{0.3\textwidth}
        \centering
        \includegraphics[width=\textwidth]{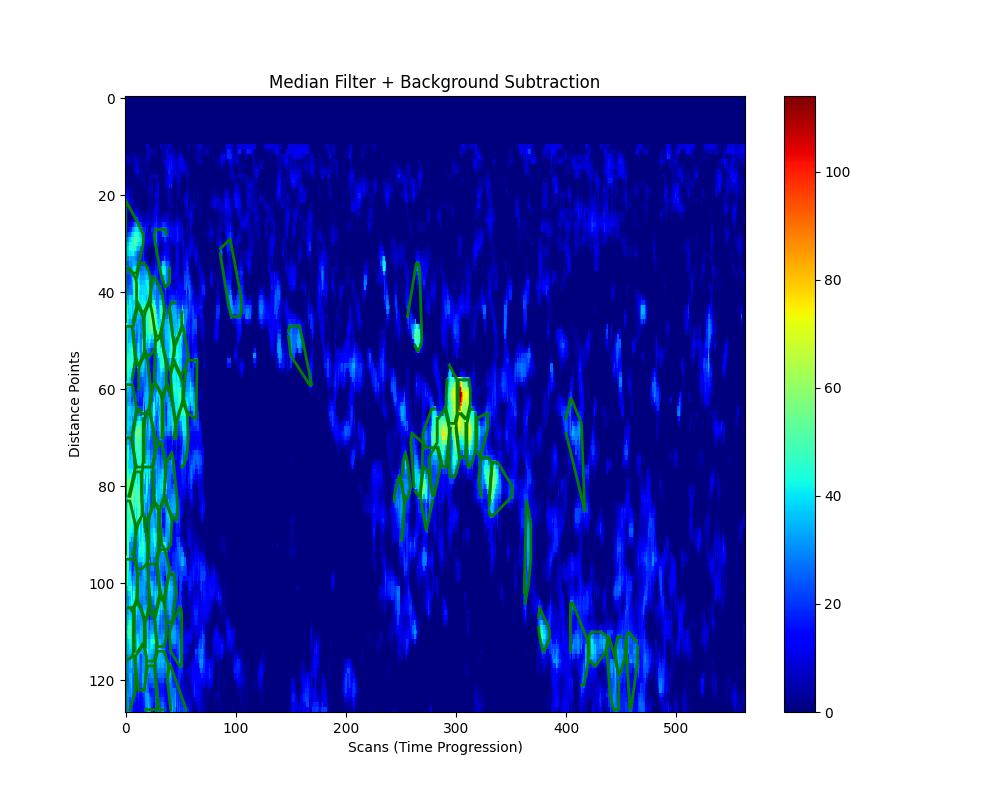}
        \caption{CSV Broom Filtering}
        \label{fig:BroomCSV}
    \end{subfigure}
    \hfill
    \begin{subfigure}[b]{0.3\textwidth}
        \centering
        \includegraphics[width=\textwidth]{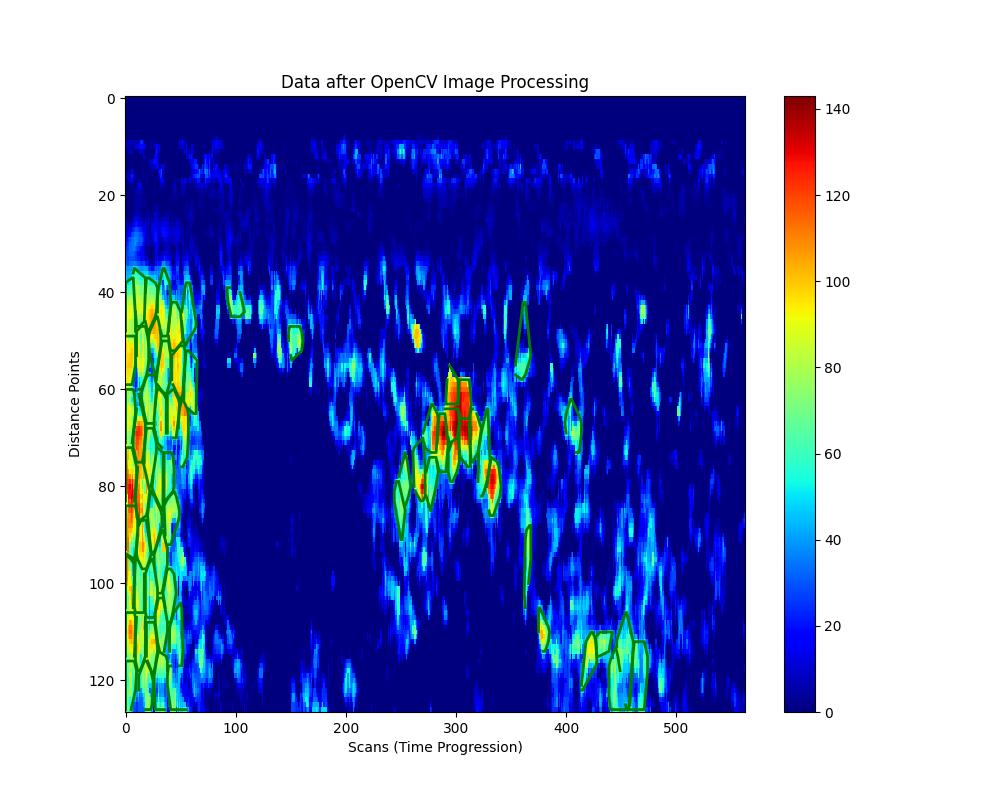}
        \caption{CV Broom Filtering}
        \label{fig:BroomCV}
    \end{subfigure}
    
    \caption{Comparison of different views of objects used in the dataset.}
    \label{fig:PolarPlotsCombined}
\end{figure*}

\subsection{Summary of Programming Workflow}
The programming implementation can be summarized through the following algorithmic steps:

\begin{enumerate}[]
    \item \textbf{Data Loading}: Import the sonar dataset using Pandas and preprocess the timestamp and scan data.
    \item \textbf{Data Cleaning}: Merge scan data to produce uniform, space-separated intensity values.
    \item \textbf{Median Filtering}: Use OpenCV's \texttt{medianBlur} to smooth the data while retaining edge information.
    \item \textbf{Background Subtraction}: Calculate and subtract the background median to isolate objects of interest.
    \item \textbf{Clustering for Object Detection}: Perform K-Means clustering on the filtered data to group similar intensity points.
    \item \textbf{Convex Hull Computation}: Compute the convex hull for each significant cluster to outline detected objects.
    \item \textbf{Evaluation}: Calculate comprehensive metrics to quantitatively assess processing quality and detection accuracy.
\end{enumerate}

%% file: Results.tex
\section{Results}
\begin{comment}
    This section presents the outcomes of the implemented sonar data processing methodologies, highlighting performance improvements in processing speed and the enhanced clarity of detected objects. A comparative analysis is conducted between the proposed Comma Separated Value (CSV) data approach and conventional Computer Vision (CV) image processing techniques applied to sonar imagery.
\end{comment}
This section presents the results of the proposed data processing methodologies. A high-level analysis is then conducted and emphasizes not only the quantitative difference in run time between the CV and the CSV approach but also the differences that may not be immediately obvious between the two results. Specifically, in situations where quantitative metrics do not represent the significance of the difference between the approaches.

\subsection{Processing Time}
\begin{comment}
    The efficiency of the proposed sonar data processing pipeline is evaluated by comparing the processing times against a standard CV-based approach. Table \ref{tab:processing_time} summarizes the filtering and total processing times for both methods.
\end{comment}

The quantitative metric of duration to run the de-noising process through one iteration of the data set is displayed in Table \ref{tab:processing_time}. As can be seen, the CSV approach previously proposed has minimal execution overhead. Alternatively, the CV image approach has both greater filtering time and processing time. This was the expected behavior, as these images are stored in formats that require more than simply reading the data as is the case with the CSV method.

\begin{table}[h]
    \centering
    \caption{Processing Time Comparison between Proposed CSV Approach and CV Image Processing}
    \begin{tabular}{@{}lll@{}}
        \toprule
        \textbf{Method} & \textbf{Filtering Time (s)} & \textbf{Total Processing Time (s)} \\ \midrule
        CSV Approach & 0.0031 & 0.0031 \\
        CV Image Processing & 0.003798 & 0.037419 \\ \bottomrule
    \end{tabular}
    \label{tab:processing_time}
\end{table}

\begin{comment}
    After running the proposed pipeline, the CSV processing method demonstrated a 91.18\% reduction in total processing time when compared to the conventional CV image processing approach. This processing time was calculated by measuring the duration from data loading to the completion of filtering. For the CV method, this also included the conversion of each data output to an image frame. Even when isolating the filtering step alone by excluding the frame production time, the proposed method remains 13.11\% faster. These improvements are primarily attributed to the elimination of computationally intensive image processing steps inherent in the CV approach, along with the use of vectorized operations on the CSV data.
\end{comment}

After running the proposed pipeline, the CSV processing method demonstrated a 91.18\% reduction in total processing time when compared to the conventional CV image processing approach. This processing time was calculated by measuring the duration from data loading to the completion of filtering. For the CV method, this also included the conversion of each data output to an image frame. Even when isolating the filtering step alone by excluding the frame production time, the proposed method remains 13.11\% faster. These improvements can be primarily attributed to the elimination of computationally intensive image processing steps inherent in the CV approach, with the  13.11\% reduction likely being due to the use of vectorized operations on the CSV data.

\subsection{Object Detection and Clarity}

\begin{comment}
    The clarity and accuracy of detected objects are critical for effective sonar data analysis. This study focuses on isolating distinct objects within the sonar scans, such as gates and submerged brooms, and evaluating the effectiveness of the proposed CSV-based filtering method compared to conventional CV-based image processing techniques.
\end{comment}

\subsubsection{Isolation and Visualization on Sonar Maps}

\begin{comment}
    The primary objective is to accurately isolate and detect objects of interest from the sonar data while minimizing background noise. Figures \ref{fig:csv_approach} and \ref{fig:cv_approach} display the processed sonar scans using the proposed CSV approach and the standard CV approach, respectively.

    \textbf{Figure \ref{fig:csv_approach}} illustrates the sonar scan after applying median filtering combined with background subtraction using the CSV approach. This method effectively reduces noise and enhances the visibility of objects, resulting in clearer and more distinct detections of the gate. In contrast, \textbf{Figure \ref{fig:cv_approach}} shows the sonar scan processed using the conventional CV image processing techniques. The CV approach introduces more artifacts and retains higher levels of background noise, which obscures object boundaries and reduces detection clarity.

\end{comment}

For effective sonar data analysis, clarity of objects is essential for object detection. As discussed in the section \textit{Data Utilized in This Study} in the Background, this study focused on isolating a gate and broom in an underwater environment to assess the effectiveness of the proposed method. As shown in Figures \ref{fig:csv_approach} and \ref{fig:cv_approach} , the output filtering of both methods varied drastically. 

For the filtering of the gate, the CV method filtered out the gate that was to be preserved in Fig. \ref{fig:cv_approach}. In contrast, the CSV method as shown in Fig. \ref{fig:csv_approach} effectively filtered out all information other than the gate.  Furthermore, the ML technique performed significantly better on the CSV data, correctly clustering the gate, whereas it was poorly clustered on the CV data.

The CV filtering proved more effective for handling the broom as shown in Fig. \ref{fig:BroomCSV}, however, it still contained a notable amount of noise after processing. In contrast, for the CSV filtering, minimal noise was contained in the final output in Fig. \ref{fig:BroomCSV}.

\subsubsection{Difference Maps}

\begin{comment}
    To quantify the improvements, difference maps were generated to compare the processed scans against the original unfiltered data.
    \textbf{Figure \ref{fig:Difference Map}} compares the difference between the processed scans and the original unfiltered data for both the CV and CSV approaches. The CSV approach (\textbf{Figure \ref{fig:CSV Difference Map}}) shows minimal residual noise and well-defined object boundaries, whereas the CV approach (\textbf{Figure \ref{fig:CV Difference Map}}) exhibits higher levels of background noise and less precise object delineation.
\end{comment}

\begin{figure}[h]
    \centering
    \begin{subfigure}[b]{0.23\textwidth}
        \centering
        \includegraphics[width=\textwidth]{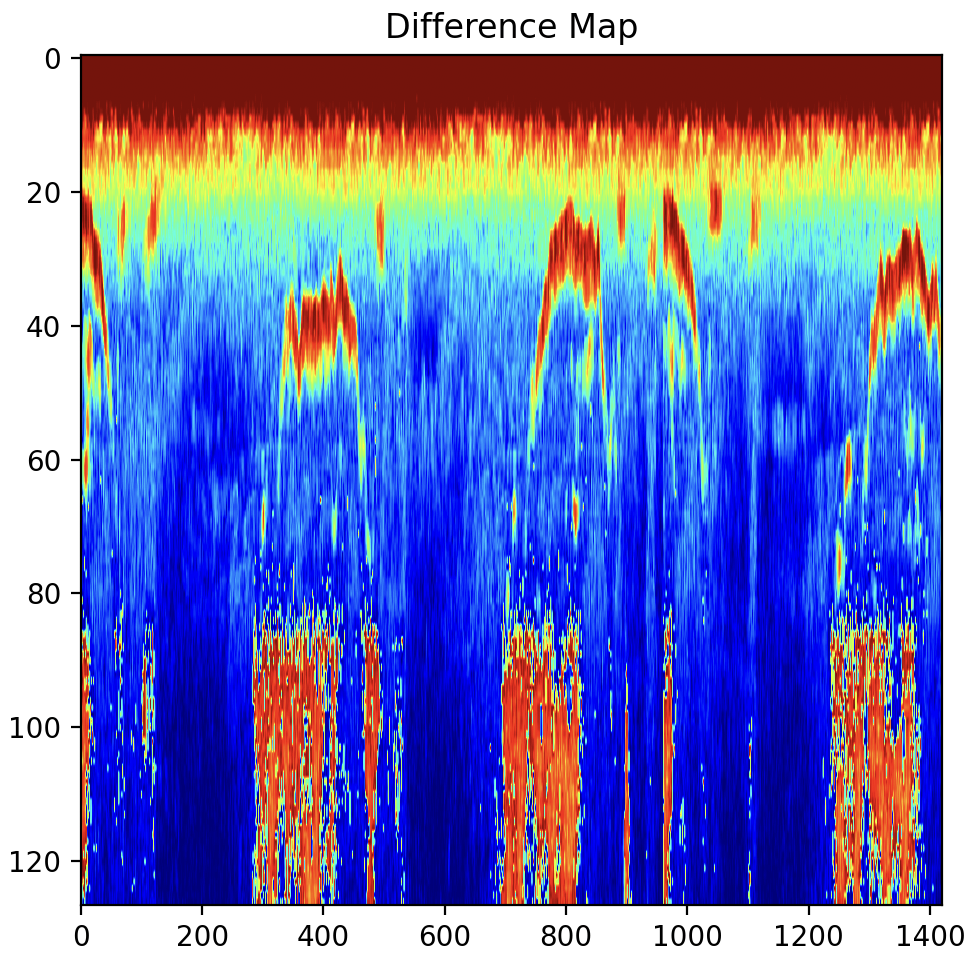}
        \caption{CV vs. Original}
        \label{fig:CV Difference Map}
    \end{subfigure}
    \hfill
    \begin{subfigure}[b]{0.23\textwidth}
        \centering
        \includegraphics[width=\textwidth]{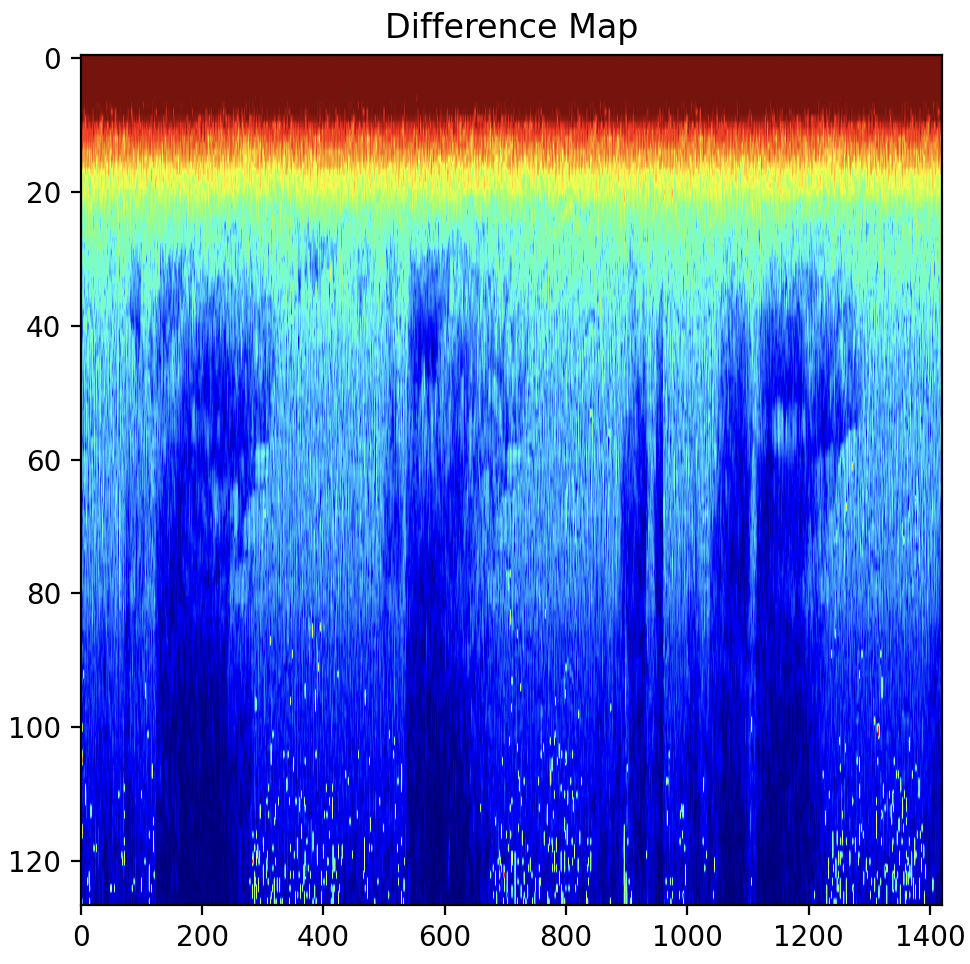}
        \caption{CSV vs. Original}
        \label{fig:CSV Difference Map}
    \end{subfigure}
    \caption{Comparison of Differences between Processed and Original Sonar Scans}
    \label{fig:Difference Map}
\end{figure}

Difference maps were created to illustrate the information removed by each method to further display the variation between them in Fig. \ref{fig:Difference Map}.  The CSV approach difference map in Fig. \ref{fig:CSV Difference Map} displays nearly all information other than the gate to be isolated having been removed, displaying a clear distinction of where the gate initially was. In contrast, the CV approach difference map in Fig. \ref{fig:CV Difference Map} includes the gate that was to be preserved, highlighting the removal of this essential feature. 

\begin{comment}
    
\begin{figure}[h]
    \centering
    \includegraphics[width=0.25\textwidth]{Images/Background.jpg}
    \caption{Background Removed Using CSV Approach}
    \label{fig:background}
\end{figure}
\end{comment}

Overall the CSV approach effectively reduced noise and enhanced the visibility of objects, resulting in clear detections of both objects. In contrast, the conventional CV image processing techniques preserved more artifacts and contained a consistently higher level of noise, obscuring objects and harming detection.

%% file: LimitationsAndFutureWork.tex
\section{Limitations and Future Work}

\subsection{Limitations}

\begin{comment}
    While the proposed method shows significant improvements in processing speed and object detection clarity, several limitations exist:
\begin{itemize}
    \item \textbf{Dataset Specificity}: Optimized for specific sonar datasets, limiting generalizability to diverse environments.
    \item \textbf{Algorithmic Constraints}: Reliance on K-Means clustering may not suit all sonar data types, especially those with irregular cluster shapes or varying densities.
    \item \textbf{Limited Sonar Types and Environments}: Currently tested on a single sonar type and environment, necessitating broader validation.
\end{itemize}
\end{comment}

Despite the significant improvements provided by this study's approach, there are still notable limitations that exist. 
This study is reliant on specific data output by a single sonar type. This data was collected by this study, and as a result, outside data is not widely explored. This is because sonar data is most commonly accessible in an image format, whereas this study set out to provide an alternative to this approach. Additionally, only one environment was available to collect data from, limiting the exploration of the proposed approach across various environments.
An additional limitation of this study was the reliance on K-Means clustering as the only ML application explored as this clustering may not apply to all sonar types. There was also no exploration of how different ML applications will perform on this data such as comparing object classification performance across data types.

\subsection{Future Work}

\begin{comment}
    To address the identified limitations and further enhance the methodology, future research will focus on the following.

\subsubsection{Expanded Dataset and Environment Diversity}
\begin{itemize}
    \item \textbf{Broader Dataset Collection}: Expand datasets to include diverse underwater scenarios, object types, and environmental conditions to enhance robustness.
    \item \textbf{Cross-Environment Validation}: Test methods across different sonar systems and environments to ensure consistent performance.
\end{itemize}

\subsubsection{Algorithmic Enhancements}
\begin{itemize}
    \item \textbf{Advanced Clustering}: Implement DBSCAN and HDBSCAN to better manage varied cluster shapes and densities, thereby improving detection accuracy.
    \item \textbf{Machine Learning Integration}: Incorporate RNNs and Transformer-based models to enhance feature extraction and object classification.
\end{itemize}
\end{comment}

To address the identified limitations and further enhance the methodology, future research will focus on the following.

Expanding the datasets to include more diverse environments and objects. Additionally, collect data from different types of sonar for verification of the proposed method across hardware.

Implementation of more advanced clustering techniques such as Hierarchical Density-Based Spatial Clustering of Applications with Noise (HDBSCAN) to better manage varied shapes and densities to further improve detection. Furthermore the development of a classification model to perform on CSV data, and a comparison of computational efficiency and accuracy to traditional CV classification models.

\begin{figure}[!htbp]
    \centering
    \includegraphics[width=0.3\textwidth]{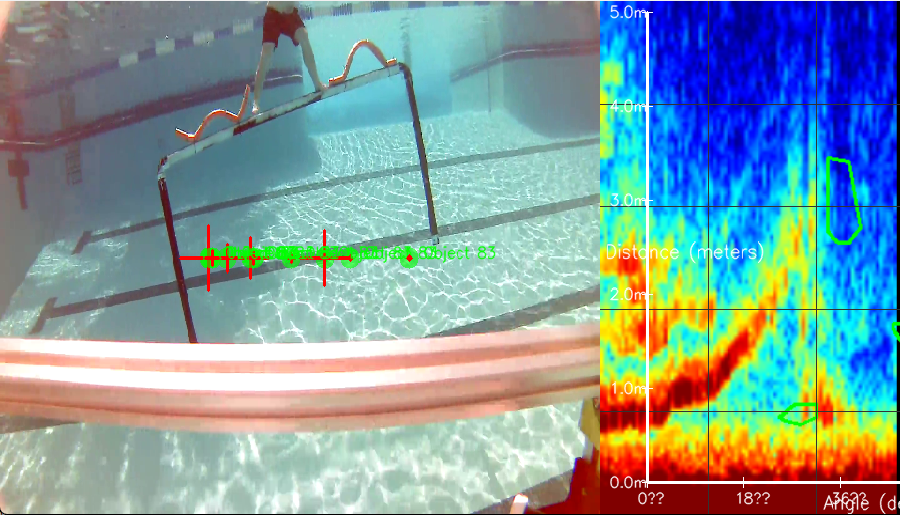}
    \caption{Experimental Sensor Fusion System on MATE ROV Submersible}
    \label{fig:sensor_fusion}
\end{figure}

Develop a sensor fusion framework combining computer vision, sonar, and DVL data for robust object tracking and classification using techniques like keypoint detection, Optical Flow, YOLO Detection, and Kalman Filtering. An initial implementation of this is shown in Fig. \ref{fig:sensor_fusion}.

%% file: Conclusion.tex
\section{Conclusion}

In this study, we proposed a novel approach for processing sonar data via CSV data, replacing the traditional image processing onboard autonomous systems. This achieved a significant reduction in computation time, with a 91.18\% reduction in total processing time when compared to traditional computer vision-based image processing, with a 13.11\% improvement in filtering speed alone. Furthermore, the use of CSV data produced cleaner, more distinct outputs which improved machine learning predictions. Future plans include the extension of hardware and environments tested on and comparison of the performance of additional machine learning methods on this data.